\documentclass[conference]{IEEEtran}
\IEEEoverridecommandlockouts
\usepackage{cite}
\usepackage{amsmath,amssymb,amsfonts}
\usepackage{algorithmic}
\usepackage{graphicx}
\usepackage{textcomp}
\usepackage{xcolor}
\usepackage{bm}
\usepackage{booktabs}       
\usepackage{amsfonts}       
\usepackage{nicefrac}       
\usepackage{microtype}      
\usepackage{xcolor}         
\usepackage{amsmath}
\usepackage{graphicx}
\usepackage{algorithm}
\usepackage{algorithmic}
\usepackage{multicol}
\usepackage{multirow}
\usepackage{stfloats}
\usepackage{bm}
\usepackage{multirow}
\usepackage{graphicx}
\usepackage{booktabs}
\usepackage{cleveref}
\usepackage{algorithm}
\usepackage{algorithmic}
\usepackage{bm}
\usepackage{amsmath}
\usepackage{amssymb}
\usepackage{bm}
\usepackage{multirow}
\usepackage{xcolor}         
\usepackage{tablefootnote}
\usepackage{threeparttable}
\usepackage{subcaption}

\def\BibTeX{{\rm B\kern-.05em{\sc i\kern-.025em b}\kern-.08em
    T\kern-.1667em\lower.7ex\hbox{E}\kern-.125emX}}

\title{FSRF: Factorization-guided Semantic Recovery for Incomplete Multimodal Sentiment Analysis}

\author{Ziyang Liu$^{1}$$\quad$
        Pengjunfei Chu$^{2}$$\quad$
        Shuming Dong$^{3}\quad$
        Chen Zhang$^{1}\quad$ 
        Mingcheng Li$^{4}\quad$ 
        Jin Wang$^{1}$\textsuperscript{*} \\
        \small$^1$School of Future Science and Engineering, Soochow University, Suzhou, China \\ 
        \small$^2$School of Advanced Manufacturing Engineering, Hefei University, Hefei, China \\
        \small$^3$College of Global Talents, BITZH, Beijing Institute of Technology, Zhuhai, Zhuhai, China\\
        \small$^4$Academy for Engineering and Technology, Fudan University, Shanghai, China \\
{\tt\small 2262403062@stu.suda.edu.cn, chupeng0426@qq.com, mingchengli21@m.fudan.edu.cn}
}

\begin{document}
\maketitle
\renewcommand{\thefootnote}{\fnsymbol{footnote}}
\footnotetext[1]{Corresponding author.} 

\begin{abstract}
In recent years, Multimodal Sentiment Analysis (MSA) has become a research hotspot that aims to utilize multimodal data for human sentiment understanding. 
Previous MSA studies have mainly focused on performing interaction and fusion on complete multimodal data, ignoring the problem of missing modalities in real-world applications due to occlusion, personal privacy constraints, and device malfunctions, resulting in low generalizability.
 To this end, we propose a Factorization-guided Semantic Recovery Framework (FSRF)  to mitigate the modality missing problem in the MSA task.
 Specifically, we propose a de-redundant homo-heterogeneous factorization module that factorizes modality into modality-homogeneous, modality-heterogeneous, and noisy representations and design elaborate constraint paradigms for representation learning.
 Furthermore, we design a distribution-aligned self-distillation module that fully recovers the missing semantics by utilizing bidirectional knowledge transfer.
 Comprehensive experiments on two datasets indicate that FSRF has a significant performance advantage over previous methods with uncertain missing modalities.
\end{abstract}


\section{Introduction}
\label{sec:intro}
Multimodal Sentiment Analysis (MSA) has become a prominent focus in recent years. 
Unlike conventional emotion recognition tasks that rely on a single modality \cite{yang2024robust}, MSA integrates various modalities to enhance the understanding of human sentiment \cite{xie2024trustworthy}.
Prior research has demonstrated that integrating complementary semantics across multiple modalities enhances the creation of more precise multimodal representations \cite{tang2022bafn}.
MSA has been well studied so far under the assumption that all modalities are available in the training and inference phases \cite{yu2021learning, yang2022emotion, li2023decoupled, yang2024asynchronous}.
However, in practical applications, modalities are often missing due to factors such as security issues, background noise, or sensor limitations. These incomplete multimodal datasets substantially impair the performance of MSA.

In recent years, many studies \cite{ lian2023gcnet, wang2023distribution, wang2020transmodality,  pham2019found, zeng2022tag,li2024unified, li2024correlation} have focused on tackling the challenge of missing modalities in MSA, which can be classified into two distinct paradigms:
(\romannumeral1) generative methods \cite{wang2023distribution, lian2023gcnet} and (\romannumeral2) joint learning methods \cite{ wang2020transmodality, li2024correlation, zeng2022tag}.
Generative methods focus on reconstructing missing features and semantics within modalities by utilizing the distributions of the accessible modalities.
For example, GCNet \cite{lian2023gcnet} proposes two graph neural network-based modules to capture speaker and temporal dependencies in conversations to address the problem of missing modality in MSA.
In contrast, joint learning methods aim to develop joint multimodal representations by exploiting correlations between modalities.
For instance, CorrKD \cite{li2024correlation} introduces a correlation-decoupled knowledge distillation framework that leverages cross-sample, cross-category, and cross-response correlations to create robust joint multimodal representations for mitigating the modality missing problem in MSA.
However, these approaches are limited by two factors:
(1) Implementing complex inter-modality interactions based on incomplete modalities ignores the extraction of task-relevant semantics and task-independent noise. (2) The recovery paradigm for missing semantics is coarse-grained and static, without sufficient consideration of high-dimensional distributional alignment.

To address the above problem, we propose a Factorization-guided Semantic Recovery Framework (FSRF) to solve the modality missing problem in MSA. Our strength comes from the following three core contributions:
(i) We propose a de-redundant homo-heterogeneous factorization module that decomposes each modality into modality-homogeneous, modality-heterogeneous, and noise representations, and design elaborate constraint paradigms for representation learning.
(ii) We propose a distribution alignment-based self-distillation module that utilizes the Sinkhorn distance and JS divergence to achieve bidirectional constraints between two networks.
(iii) Extensive experiments conducted on two datasets demonstrate that FSRF has a significant performance advantage over previous methods with uncertain missing modalities and comparable performance with complete modalities.

\section{Methodology}
\subsection{Problem Formulation}
Given a multimodal video dataset $\mathcal{D} = \{\bm{X}_i, \bm{Y}_i\}^N_{i=1}$, where $N$ is the number of samples.
Each sample is containing three distinct modalities, denoted as $\bm{S}=[\bm{M}_L, \bm{M}_A, \bm{M}_V]$, where $\bm{M}_L\in \mathbb{R}^{T_L\times d_L}, \bm{M}_A \in \mathbb{R}^{T_A\times d_A}$, and  $\bm{M}_V\in\mathbb{R}^{T_V\times d_V}$ denote language, audio, and visual modalities, respectively. 
$\Phi=\{L,A,V\}$ denotes the set of modality types.
$T_{m}(\cdot)$ and $d_{m}(\cdot)$ represent the sequence length and the embedding dimension, where $m \in \Phi$.
To effectively replicate the uncertainty of modality missingness observed in real-world scenarios, we define two types of missing cases:
(1) \textit{intra-modality missingness}, which indicates some frame-level features in the modality sequences are missing. (2) \textit{inter-modality missingness}, which indicates some modalities are entirely missing.
The goal is to perform utterance-level sentiment recognition by utilizing multimodal data that includes missing modalities.

\begin{figure*}[t]
  \centering
  \includegraphics[width=0.75\textwidth]{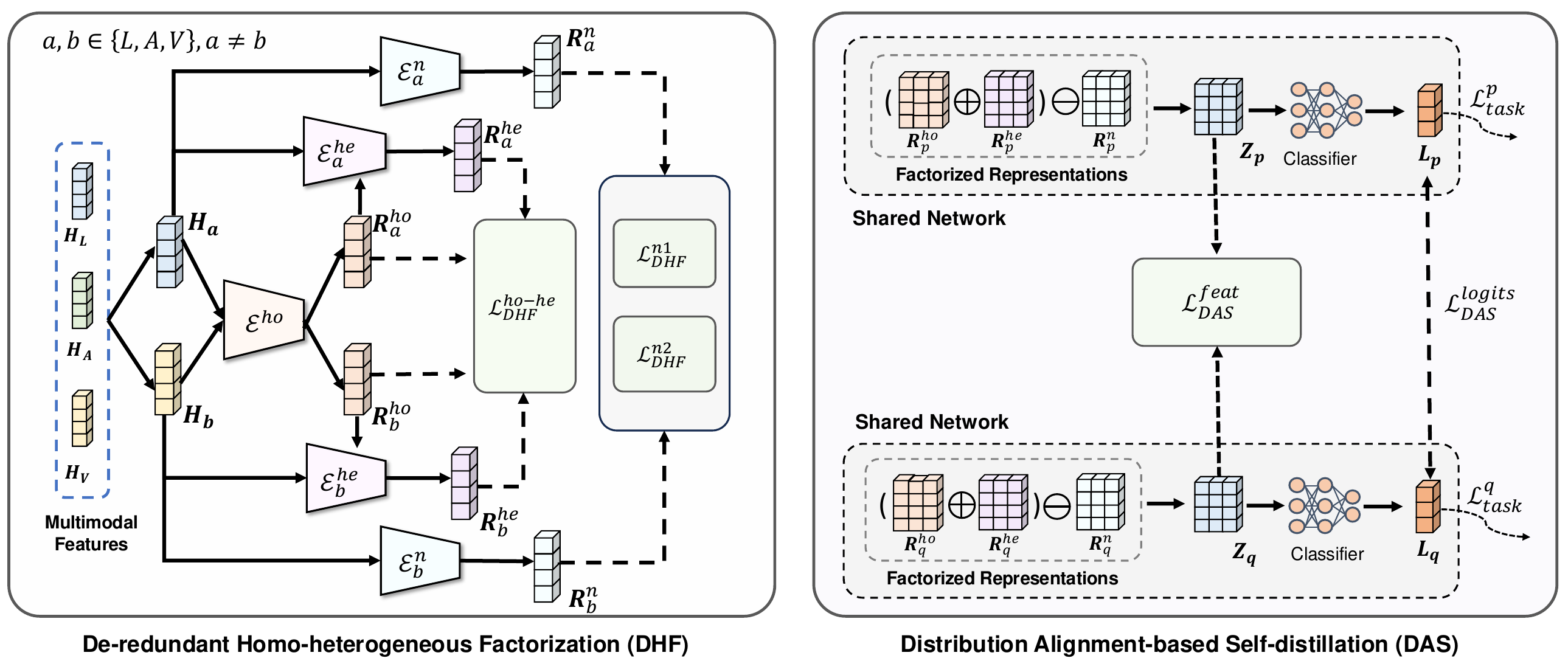}
  \caption{The overall architecture of FSRF.  We propose a De-redundant Homo-Heterogeneous Factorization (DHF) mechanism and a Distributional Alignment-based Self-Distillation (DAS) module to accurately recover missing information.}
  \label{fig:overall_framework1}
  \vspace{-5pt}
\end{figure*}

\subsection{Modality Preprocessing}
The Modality Random Missing (MRM) strategy performs both intra-modality missing and inter-modality missing, generating two heterogeneous modality missing samples $\bm{S}_p$ and $\bm{S}_q$ from the input complete modality sample $\bm{S}$.
Our framework receives  $\bm{S}_p$ and $\bm{S}_q$ as input in parallel and symmetrically. 
For brevity, here we denote a particular sample of the input to the network with $\bm{S}=[\bm{M}_L, \bm{M}_A, \bm{M}_V]$.
The language modality is fed into the pre-trained BERT
\cite{devlin2018bert} to obtain the average feature for each word, denoted as $\bm{C}_L \in \mathbb{R}^{d_L}$. 
The audio and visual modalities plus the positional encoding are then fed into the Transformer encoder ~\cite{vaswani2017attention}, whose last elements of the output are expressed as $\bm{C}_A \in \mathbb{R}^{d_A}$ and $\bm{C}_V \in \mathbb{R}^{d_V}$.
Each $\bm{C}_{m}$  is passed through a fully-connected layer to unify the dimensions of the features to obtain $\bm{H}_m$.
\subsection{De-redundant Homo-Heterogeneous Factorization}
Modality missing destroys the contextual semantics inherent in multimodal data, significantly impacting the effectiveness of multimodal fusion and semantic extraction.
Previous studies of MSA with missing modalities ~\cite{yuan2021transformer, wang2020transmodality, pham2019found} focus on designing complex cross-modal interaction paradigms to mine valuable information in incomplete modalities.
These approaches treat incomplete modalities as the minimal interaction unit and ignore the shared, specific, and noisy information contained in the modalities, thus introducing significant cumulative errors.
Therefore, we propose a De-redundant Homo-heterogeneous Factorization (DHF) module, to fully exploit the multilevel representations in modality.

Specifically, the DHF factorizes each modality into three categories of representations: 
(1) Modality-homogeneous representation, denoting the representation reflecting the sentiment homogeneity shared among all modalities. 
(2) Modality-heterogeneous representation, which represents different modality-specific information for characterizing sentiment in different modalities. 
(3) Noise representation, which indicates the noise information contained in each modality.
For each modality $m \in \Phi$, we obtain the modality-homogeneous representation by a modality-shared encoder, denoted as $\bm{R}_m^{ho} = \mathcal{E}^{ho}(\bm{{H}_m})$, a modality-heterogeneous representation by a modality-specific encoder, denoted as  $\bm{R}_m^{he} = \mathcal{E}_m^{he}(\bm{{H}}_m,\bm{R}_m^{ho} )$, and a noise representation by a modality-specific encoder, denoted as  $\bm{R}_m^{n} = \mathcal{E}_m^{n}(\bm{{H}}_m)$.
All encoders are composed of multi-layer perceptrons with the ReLU activation.

Modality-homogeneous representations should contain sentiment information shared among all modalities in the sample, and modality-heterogeneous representations of different modalities should be inconsistent. To achieve this objective, we propose a contrastive learning-based representation constraint mechanism, which employs NT-Xent loss \cite{chen2020simple} to constrain the distribution of different representations in the latent space.
It has stronger constraints and training stability than traditional contrastive learning loss through normalization and temperature parameters. 
The loss is denoted as:
\begin{equation}
    \mathcal{L}_{DHF}^{ho-he}= -\frac{1}{6}\sum_{a \in \Phi} \,\sum_{b \in \Phi, a\neq b} log \frac{\mathcal{G}(\bm{R}^{ho}_a, \bm{R}^{ho}_b)}{\mathcal{G}{(\bm{R}^{ho}_a, \bm{R}^{ho}_b)}+ \mathcal{G}(\bm{R}^{ho}_a, \bm{R}^{he}_a)},
\end{equation}
where $\mathcal{G}(\bm{X}, \bm{Y}) = exp (-\mathcal{D}(\bm{X}, \bm{Y}) / \tau)$, $\mathcal{D}(\bm{X}, \bm{Y})$ is the cosine distance of $\bm{X}$ and $\bm{Y}$, and $\tau$ is the temperature hyper-parameter.
Additionally, each modality has its unique noise representation while remaining consistent across samples of the same modality, which is constrained as follows:
\begin{equation}
\begin{aligned}
    \small
    \mathcal{L}_{DHF}^{n_1} &= \frac{1}{3\,N_B\,(N_B-1)}\sum_{i=1}^{N_B} \, \sum_{j=1,j \neq i}^{N_B} \, \sum_{m \in \Phi} \mathcal{D} \left(\bm{R}^n_{i,m}, \bm{R}^n_{j,m} \right) \\
    & 
    +\frac{1}{6\,N_B}\sum_{i=1}^{N_B}  \, \sum_{a \in \Phi} \, \sum_{b \in \Phi, b\neq a} \max \left(\epsilon-\mathcal{D} \left(\bm{R}^n_{i,a}, \bm{R}^n_{i,b}\right), 0\right),
\end{aligned}
\end{equation}
where $N_B$ is the number of samples in a mini-batch and $\epsilon$ is the predefined distance margin.
Moreover, to enhance the compactness of the noise representation in low-dimensional space to facilitate the stable extraction of noisy information, we minimize the entropy of the noise representation. 
 Simultaneously, we use regularization to control the noise representation to avoid it being excessively compact and losing diversity.
 This regularized entropy loss is expressed as:
\begin{equation}
\footnotesize
    \mathcal{L}_{DHF}^{n_2}= \frac{1}{3}\sum_{m\in \Phi}\left( \mathcal{H}\left(\bm{R}^n_m\right)+ (\left\|\bm{R}^n_m-\mu_m\right\|^2 + \left\|Var(\bm{R}^n_m) - \sigma_m^2)\right\|^2\right),
\end{equation}
where $\mathcal{H}(\cdot)=\frac{1}{2} \log \left((2 \pi e)^d \operatorname{det}(\Sigma)\right)$ is the entropy formula for the multi-dimensional Gaussian distribution, and $\Sigma$ is the covariance matrix, and $\mu_m$ and $\sigma_{m}^2$ are the predefined mean and variance.
%
Therefore, the loss of DHF is expressed as:
\begin{equation}
    \mathcal{L}_{DHF} = \mathcal{L}_{DHF}^{ho-he} + \mathcal{L}_{DHF}^{n_1} + \mathcal{L}_{DHF}^{n_2}.
\end{equation}
\subsection{Distribution Alignment-based Self-Distillation}
Conventional knowledge distillation techniques for handling missing modalities utilize complete-modality teacher networks to direct the training of missing-modality student networks.
These methods are hindered by several drawbacks, such as the necessity for high-performing teacher networks, substantial training expenditures, and the static and coarse-grained information transfer \cite{ cho2021dealing, hu2020knowledge}. 
To tackle the aforementioned challenges, inspired by \cite{li2024unified}, we propose a Distribution Alignment-based Self-distillation (DAS) module to progressively learn the representation consistency and recover the missing sentiment semantics through a hierarchical self-distillation paradigm. 
 Specifically, DAS performs bi-directional information delivery within a unified and shared network to achieve feature and logits consistency constraints between two heterogeneous modality missing samples.
This learning paradigm reduces the unidirectional dependence on knowledge~\cite{morcos2018importance} and provides two key advantages: transferring knowledge from richer modalities to those with fewer modalities helps recover information lost due to missing modalities, while the reverse flow of knowledge refines and strengthens useful features. In conclusion, DAS enables the model to generate more robust joint multimodal representations.

\noindent\textbf{Feature-level Distillation}. 
 We add all the modality-homogeneous representations and modality-heterogeneous representations and subtract all the noisy representations to obtain the joint multimodal representations $\bm{Z}_p$ and $\bm{Z}_q$ of the two heterogeneous samples.
To accurately measure the holistic discrepancy between the distributions of the two representations, we utilize the Sinkhorn distance \cite{cuturi2013sinkhorn}. 
It is a variant of the classical optimal transport (OT) distance \cite{villani2009optimal}, which measures the minimal cost of transporting mass between two distributions.
Unlike the traditional Wasserstein distance \cite{villani2009optimal}, Sinkhorn distance introduces a regularization term, which makes the computation more efficient and helps avoid numerical instability issues.
Mathematically, given two probability distributions $\mu$ and $\nu$ the Sinkhorn distance is defined as:
\begin{equation}
    S_\epsilon(\mu, \nu)=\min _{\gamma \in \Pi(\mu, \nu)}\left(\int_{\mathcal{X} \times \mathcal{Y}} c(x, y) d \gamma(x, y)+\epsilon H(\gamma)\right),
\end{equation}
where $\Pi(\mu, \nu)$ is the set of joint distributions that satisfy the marginal constraints, and $c(x, y)=\|x-y\|^2$ is the cost function. $H(\gamma)$ is the entropy of the joint distribution $\gamma$, defined as:
\begin{equation}
    H(\gamma)=-\int_{\mathcal{X} \times \mathcal{Y}} \gamma(x, y) \log (\gamma(x, y)) \,d x \, d y,
\end{equation}
The regularization parameter $\epsilon$ controls the strength of the entropy term, balancing the transportation cost and the entropy. 
Therefore, we adopt Sinkhorn distance to measure the discrepancy between $\bm{Z}_p$ and $\bm{Z}_q$, which is represented as:
\begin{equation}
\label{eq:feature_loss}
\mathcal{L}_{DAS}^{feat} = S_\epsilon(\bm{Z}_p,\bm{Z}_q),
\end{equation}

\noindent\textbf{Logits-level Distillation}.
To further recover the missing sentiment semantics, we constrain the logits of the two networks.
The representations $\bm{Z}_p$ and $\bm{Z}_q$ pass through fully connected layers and softmax layer to get logits $\bm{L}_p$ and $\bm{L}_q$.
The Jensen-Shannon (JS) divergence is used as the measure of discrepancy, which serves as a symmetrical metric for assessing the similarity between two probability distributions, expressed as:
\begin{equation}
\small
\label{eq:logits_loss}
\begin{aligned}
    \mathcal{L}_{DAS}^{logits} =\mathcal{D}_{JS}(\bm{L}_p||\bm{L}_q)) =\frac{1}{2}(\mathcal{D}_{KL}(\bm{L}_p|\bm{M})) + \mathcal{D}_{KL}(\bm{L}_q || \bm{M})),
\end{aligned}
\end{equation}
where $\bm{M}$ is the average distribution of $\bm{L}_p$ and $\bm{L}_q$.
Ultimately, the loss of DAS is denoted as:
\begin{equation}
    \mathcal{L}_{DAS} = \mathcal{L}_{DAS}^{feat} + \mathcal{L}_{DAS}^{logits}.
\end{equation}
$\bm{L}_p$ and $\bm{L}_q$ are used to calculate the task loss $\mathcal{L}_{task}$.
In regression and classification tasks. The task losses are Mean Squared Error (MSE) and cross-entropy losses, respectively. Eventually, the overall optimization objective $\mathcal{L}_{total}$ is expressed as $\mathcal{L}_{total} = \mathcal{L}_{task} + \lambda_1 \mathcal{L}_{DHF} + \lambda_2 \mathcal{L}_{DAS}$.
where the hyperparameters $\lambda_1$ and $\lambda_2$ are 0.2 and 0.1.

\section{Experiments}
\subsection{Datasets and Implementation Details}

\noindent\textbf{Datasets.} The MOSI \cite{zadeh2016mosi} dataset is a widely used benchmark for MSA, comprising 2,199 opinion video clips. It is divided into 1,284 clips for training, 229 for validation, and 686 for testing. 
MOSEI \cite{zadeh2018multimodal}, an MSA dataset with 23,454 movie video clips, which is split into 16,326 training, 1,871 validation, and 4,659 testing samples.
Each instance in both MOSI and MOSEI is assigned a sentiment label ranging from -3 (strongly negative) to +3 (strongly positive). 
For our evaluations, we compute the F1 score for positive/negative classification on these datasets.
The language modality is encoded into 768-dimensional vectors using the pre-trained BERT~\cite{devlin2018bert}.
We utilize the COVAREP toolkit~\cite{degottex2014covarep} to extract 74-dimensional acoustic features for the audio modality, which include 12 Mel-frequency cepstral coefficients, voiced/unvoiced segmentation features, and glottal source parameters.
For the visual modality, we employ the Facet toolkit~\cite{baltruvsaitis2016openface} to extract 35 facial action units that capture various facial movements.

\noindent\textbf{Implementation details.} 
Regarding the MOSI \cite{zadeh2016mosi} and MOSEI \cite{zadeh2018multimodal} datasets, we use the aligned multimodal sequences therein as the original input for the FSRF.
All models are implemented using the Pytorch \cite{paszke2017automatic} toolbox with four NVIDIA Tesla A800 GPUs. 
The Adam optimizer \cite{kingma2014adam} is employed for network optimization.
For the MOSI and MOSEI datasets, the hyper-parameters are set as follows: the learning rates are $\{1e-4, 2e-4\}$, the batch sizes are $\{16, 32\}$, the epoch numbers are $\{20, 30\}$. The embedding dimension is $128$ on all two datasets. 
The missing features are replaced with zero values.
To ensure fairness in our comparisons, we have reproduced several State-Of-The-Art (SOTA) methods and integrated them into our modality missing experimental setup.
All experimental results are the average of the 10 random seed cases.

\subsection{Comparison with State-of-the-art Methods}

We select eight representative SOTA methods as baselines for comparison with FSRF. 
Specifically, to demonstrate the advantage of FSRF in the case of missing modality, we choose five missing-modality methods, including (1) joint learning methods, \emph{i.e.}, MCTN \cite{pham2019found}, TransM \cite{ wang2020transmodality}, and CorrKD \cite{li2024correlation} and (2) generative methods \emph{i.e.}, SMIL \cite{ma2021smil} and GCNet \cite{lian2023gcnet}. 
Additionally, to perform a more comprehensive comparison, we also select five complete-modality methods: Self-MM \cite{yu2021learning}, CubeMLP \cite{sun2022cubemlp}, and DMD \cite{li2023decoupled}.
We analyze in detail the performance of the proposed FSRF and baseline models in the two modality-missing cases.

\begin{table*}[t]
\centering
\caption{Comparison results under inter-modality missingness cases on MOSI and MOSEI.}
\renewcommand{\arraystretch}{1.1}
\setlength{\tabcolsep}{8pt}
\label{comp_inter_1}
\resizebox{\textwidth}{!}{%
\begin{threeparttable}[b]
\begin{tabular}{c|c|cccccccc}
\toprule
\multirow{2}{*}{Datasets} & \multirow{2}{*}{Models} & \multicolumn{8}{c}{Testing Conditions}                                                                                                                                                                                                                                                   \\ \cmidrule{3-10} 
                                   &                         & \multicolumn{1}{c|}{\{$l$\}}          & \multicolumn{1}{c|}{\{$a$\}}          & \multicolumn{1}{c|}{\{$v$\}}          & \multicolumn{1}{c|}{\{$l,a$\}}        & \multicolumn{1}{c|}{\{$l,v$\}}        & \multicolumn{1}{c|}{\{$a,v$\}}        & \multicolumn{1}{c|}{Avg. (Missing)}           & \{$l,a,v$\}      \\ \midrule
\multirow{9}{*}{MOSI}              & Self-MM \cite{yu2021learning}$^\S$
                & \multicolumn{1}{c|}{67.80}          & \multicolumn{1}{c|}{40.95}          & \multicolumn{1}{c|}{38.52}          & \multicolumn{1}{c|}{69.81}          & \multicolumn{1}{c|}{74.97}          & \multicolumn{1}{c|}{47.12}          & \multicolumn{1}{c|}{56.53}          & \textbf{84.64} \\
                                   & CubeMLP  \cite{sun2022cubemlp}$^\S$
                & \multicolumn{1}{c|}{64.15}          & \multicolumn{1}{c|}{38.91}          & \multicolumn{1}{c|}{43.24}          & \multicolumn{1}{c|}{63.76}          & \multicolumn{1}{c|}{65.12}          & \multicolumn{1}{c|}{47.92}          & \multicolumn{1}{c|}{53.85}          & 84.57          \\
                                   & DMD \cite{li2023decoupled}$^\S$ 
                    & \multicolumn{1}{c|}{68.97}          & \multicolumn{1}{c|}{43.33}          & \multicolumn{1}{c|}{42.26}          & \multicolumn{1}{c|}{70.51}          & \multicolumn{1}{c|}{68.45}          & \multicolumn{1}{c|}{50.47}          & \multicolumn{1}{c|}{57.33}          & 84.50          \\
                                   & MCTN \cite{pham2019found}$^\ddag$
                   & \multicolumn{1}{c|}{75.21}          & \multicolumn{1}{c|}{59.25}          & \multicolumn{1}{c|}{58.57}          & \multicolumn{1}{c|}{77.81}          & \multicolumn{1}{c|}{74.82}          & \multicolumn{1}{c|}{64.21}          & \multicolumn{1}{c|}{68.31}          & 80.12          \\
                                   & TransM \cite{wang2020transmodality}$^\ddag$
                 & \multicolumn{1}{c|}{77.64}          & \multicolumn{1}{c|}{63.57}          & \multicolumn{1}{c|}{56.48}          & \multicolumn{1}{c|}{82.07}          & \multicolumn{1}{c|}{80.90}          & \multicolumn{1}{c|}{67.24}          & \multicolumn{1}{c|}{71.32}          & 82.57          \\
                                   & SMIL \cite{ma2021smil}$^\ddag$
                   & \multicolumn{1}{c|}{78.26}          & \multicolumn{1}{c|}{67.69}          & \multicolumn{1}{c|}{59.67}          & \multicolumn{1}{c|}{79.82}          & \multicolumn{1}{c|}{79.15}          & \multicolumn{1}{c|}{71.24}          & \multicolumn{1}{c|}{72.64}          & 82.85          \\
                                   & GCNet \cite{lian2023gcnet}$^\ddag$
                  & \multicolumn{1}{c|}{80.91}          & \multicolumn{1}{c|}{65.07}          & \multicolumn{1}{c|}{58.70}          & \multicolumn{1}{c|}{\textbf{84.73}} & \multicolumn{1}{c|}{83.58} & \multicolumn{1}{c|}{70.02}          & \multicolumn{1}{c|}{73.84}          & 83.20          \\
                                   & CorrKD \cite{li2024correlation}$^\ddag$                & \multicolumn{1}{c|}{81.20}          & \multicolumn{1}{c|}{66.52}          & \multicolumn{1}{c|}{60.72}          & \multicolumn{1}{c|}{83.56}          & \multicolumn{1}{c|}{82.41}          & \multicolumn{1}{c|}{73.74}          & \multicolumn{1}{c|}{74.69}          & 83.94          \\
                                   & \textbf{FSRF (Ours)$^\ddag$}           & \multicolumn{1}{c|}{\textbf{\,\,\,82.58$^*$}} & \multicolumn{1}{c|}{\textbf{\,\,\,69.23$^*$}} & \multicolumn{1}{c|}{\textbf{\,\,\,64.68$^*$}} & \multicolumn{1}{c|}{\,\,\,84.19$^*$}          & \multicolumn{1}{c|}{\,\,\,\textbf{83.72}$^*$}          & \multicolumn{1}{c|}{\,\,\,\textbf{76.25$^*$}} & \multicolumn{1}{c|}{\,\,\,\textbf{76.78$^*$}} & \,\,\,84.05$^*$          \\ \midrule
\multirow{9}{*}{MOSEI}             & Self-MM \cite{yu2021learning}$^\S$
                & \multicolumn{1}{c|}{71.53}          & \multicolumn{1}{c|}{43.57}          & \multicolumn{1}{c|}{37.61}          & \multicolumn{1}{c|}{75.91}          & \multicolumn{1}{c|}{74.62}          & \multicolumn{1}{c|}{49.52}          & \multicolumn{1}{c|}{58.79}          & 83.69          \\
                                   & CubeMLP  \cite{sun2022cubemlp}$^\S$                & \multicolumn{1}{c|}{67.52}          & \multicolumn{1}{c|}{39.54}          & \multicolumn{1}{c|}{32.58}          & \multicolumn{1}{c|}{71.69}          & \multicolumn{1}{c|}{70.06}          & \multicolumn{1}{c|}{48.54}          & \multicolumn{1}{c|}{54.99}          & 83.17          \\
                                   & DMD \cite{li2023decoupled}$^\S$                    & \multicolumn{1}{c|}{70.26}          & \multicolumn{1}{c|}{46.18}          & \multicolumn{1}{c|}{39.84}          & \multicolumn{1}{c|}{74.78}          & \multicolumn{1}{c|}{72.45}          & \multicolumn{1}{c|}{52.70}          & \multicolumn{1}{c|}{59.37}          & \textbf{84.78} \\
                                   & MCTN \cite{pham2019found}$^\ddag$                   & \multicolumn{1}{c|}{75.50}          & \multicolumn{1}{c|}{62.72}          & \multicolumn{1}{c|}{59.46}          & \multicolumn{1}{c|}{76.64}          & \multicolumn{1}{c|}{77.13}          & \multicolumn{1}{c|}{64.84}          & \multicolumn{1}{c|}{69.38}          & 81.75          \\
                                   & TransM \cite{wang2020transmodality}$^\ddag$
                 & \multicolumn{1}{c|}{77.98}          & \multicolumn{1}{c|}{63.68}          & \multicolumn{1}{c|}{58.67}          & \multicolumn{1}{c|}{80.46}          & \multicolumn{1}{c|}{78.61}          & \multicolumn{1}{c|}{62.24}          & \multicolumn{1}{c|}{70.27}          & 81.48          \\
                                   & SMIL \cite{ma2021smil}$^\ddag$
                   & \multicolumn{1}{c|}{76.57}          & \multicolumn{1}{c|}{65.96}          & \multicolumn{1}{c|}{60.57}          & \multicolumn{1}{c|}{77.68}          & \multicolumn{1}{c|}{76.24}          & \multicolumn{1}{c|}{66.87}          & \multicolumn{1}{c|}{70.65}          & 80.74          \\
                                   & GCNet \cite{lian2023gcnet}$^\ddag$
                  & \multicolumn{1}{c|}{80.52}          & \multicolumn{1}{c|}{66.54}          & \multicolumn{1}{c|}{61.83}          & \multicolumn{1}{c|}{81.96}          & \multicolumn{1}{c|}{81.15}          & \multicolumn{1}{c|}{69.21}          & \multicolumn{1}{c|}{73.54}          & 82.35          \\
                                   & CorrKD \cite{li2024correlation}$^\ddag$ 
                 & \multicolumn{1}{c|}{80.76}          & \multicolumn{1}{c|}{66.09}          & \multicolumn{1}{c|}{62.30}          & \multicolumn{1}{c|}{81.74}          & \multicolumn{1}{c|}{81.28}          & \multicolumn{1}{c|}{71.92}          & \multicolumn{1}{c|}{74.02}          & 82.16          \\
                                   & \textbf{FSRF (Ours)$^\ddag$}          & \multicolumn{1}{c|}{\,\,\,\textbf{82.62$^*$}} & \multicolumn{1}{c|}{\,\,\,\textbf{69.12$^*$}} & \multicolumn{1}{c|}{\,\,\,\textbf{65.04$^*$}} & \multicolumn{1}{c|}{\,\,\,\textbf{82.27$^*$}} & \multicolumn{1}{c|}{\,\,\,\textbf{81.35$^*$}} & \multicolumn{1}{c|}{\,\,\,\textbf{74.98$^*$}} & \multicolumn{1}{c|}{\,\,\,\textbf{75.90}$^*$} & \,\,\,83.14$^*$          \\ \bottomrule
\end{tabular}%
\begin{tablenotes}
\item $^\S$ means the complete-modality methods and $^\ddag$ means the missing-modality methods. 
T-test is conducted and $^*$ indicates that $p < 0.05$  (compared with CorrKD).
\end{tablenotes}
\end{threeparttable}
}
\end{table*}

\noindent \textbf{Analysis of inter-modality missingness.}
To simulate testing conditions of inter-modality missingness, we remove certain entire modalities from the samples. Tables \ref{comp_inter_1}  contrast the models' resilience to inter-modality missingness.
We define ``$\{l\}$'' as the testing condition for both audio and visual modality are missing and 
``$\{l, a, v\}$'' as  the testing condition for the complete modality.
``Avg. (Missing)’’ represents the mean performance of six testing conditions.
We have the following key findings:
(\romannumeral1)  
The inter-modality missingness leads to a decline in performance for all models, highlighting that combining complementary information from different modalities strengthens the sentiment semantics in joint representations.
(\romannumeral2)  Under conditions where modalities are missing, the performance decline of methods using complete modalities is significantly greater than that of methods with missing modalities. This suggests that the missing-modality methods leverage their semantic-recovery learning approach to partially alleviate the impact of missing modalities.
(\romannumeral3) Among all the methods, our FSRF has the best performance, proving its strong robustness. 
For instance, on the MOSI dataset, FSRF's average F1 score is improved by $2.09\%$ compared to CorrKD, and in particular by $3.92\%$ in the testing condition $\{v\}$.
The strength of this approach lies in its capacity to perform fine-grained representation factorization and execute precise bidirectional supervision.
(iv) Different modalities have varying impacts on sentiment analysis. In unimodal situations, methods relying solely on the language modality achieve the best performance, with results similar to those obtained using all modalities. In bimodal cases, combinations involving the language modality outperform others, indicating that the language modality’s rich, high-level semantics are crucial for recovering missing information.

\begin{figure}[t]
  \centering
     \setlength{\belowcaptionskip}{0pt}
  \includegraphics[width=1\linewidth]{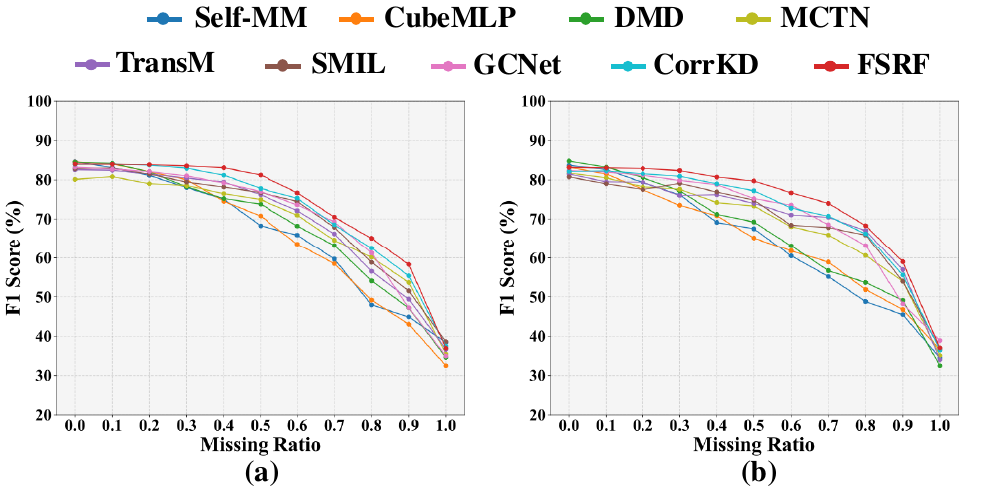}
    \vspace{-3pt}
  \caption{ 
  Comparison results under intra-modality missingness cases on (a) MOSI and (b) MOSEI. 
 }
  \label{comp_intra_1}
  \vspace{-3pt}
\end{figure}

\noindent \textbf{Analysis of intra-modality missingness.}
Intra-modality missingness is simulated by randomly discarding frame-level features from sequences at varying rates, with the missing ratio  $p \in \{0.1, 0.2, \cdots, 1.0\}$.
The performance curves of models with different $p$ values are illustrated in Figure \ref{comp_intra_1}, offering an intuitive depiction of the robustness of all models. 
These results yield several key observations.
(\romannumeral1)
As the ratio $p$ grows, every model’s performance deteriorates. The finding suggests that intra-modality missingness induces considerable sentiment semantic loss and undermines the coherence of the integrated multimodal representations.
(\romannumeral2)
In comparison to complete-modality methods, our FSRF demonstrates notable performance improvements under missing-modality conditions and achieves results that are competitive with those of complete modalities.
(\romannumeral3) Contrary to missing-modality methods, our FSRF demonstrates enhanced robustness.
By factorizing modalities and applying high-dimensional distribution constraints within the self-distillation module, FSRF effectively recovers missing features and produces robust multimodal representations.

\begin{table}[t]
  \centering
  \renewcommand{\arraystretch}{0.8}
\setlength{\tabcolsep}{8pt}
    \caption{Ablation results of FSRF on two datasets. ``Intra-MM'', ``Inter-MM'', and ``CM'': intra-modality missingness, inter-modality missingness, and complete modality cases.}
    \label{table_ablation_FSRF}
  \begin{subtable}{0.5\textwidth}
    \caption{Ablation results on the MOSI dataset.}
    \centering
    \resizebox{0.8\textwidth}{!}{%
   \begin{tabular}{cccc}
\toprule
\multirow{3}{*}{Models} & \multicolumn{3}{c}{Testing Conditions}           \\ \cmidrule{2-4} 
                        & Intra-MM       & Inter-MM       & CM             \\
                        & Avg. F1 $\uparrow$      & Avg. F1 $\uparrow$       & F1 $\uparrow$             \\ \midrule
\textbf{FSRF}           & \textbf{73.39} & \textbf{76.78} & \textbf{84.05} \\
w/o DHF                 & 70.94          & 73.31          & 82.92          \\
w/o DAS                 & 71.36          & 74.83          & 83.11          \\ \bottomrule
\end{tabular}%
}
  \end{subtable}
  
  \hfill
  
  \begin{subtable}{0.5\textwidth}
  \caption{Ablation results on the MOSEI dataset.}
    \centering
    \resizebox{0.8\textwidth}{!}{%
   \begin{tabular}{cccc}
\toprule
\multirow{3}{*}{Models} & \multicolumn{3}{c}{Testing Conditions}           \\ \cmidrule{2-4} 
                        & Intra-MM       & Inter-MM       & CM             \\
                        & Avg. F1 $\uparrow$       & Avg. F1 $\uparrow$       & F1  $\uparrow$            \\ \midrule
\textbf{FSRF}           & \textbf{72.94} & \textbf{75.90} & \textbf{83.14} \\
w/o DHF                 & 70.43          & 73.81          & 82.51          \\
w/o DAS                 & 71.06          & 74.33          & 82.27          \\ \bottomrule
\end{tabular}%
}
  \end{subtable}
  \vspace{-6pt}

\end{table}

\subsection{Ablation Studies}
\begin{figure}[t]
\vspace{-4pt}
  \centering
  \includegraphics[width=0.9\linewidth]{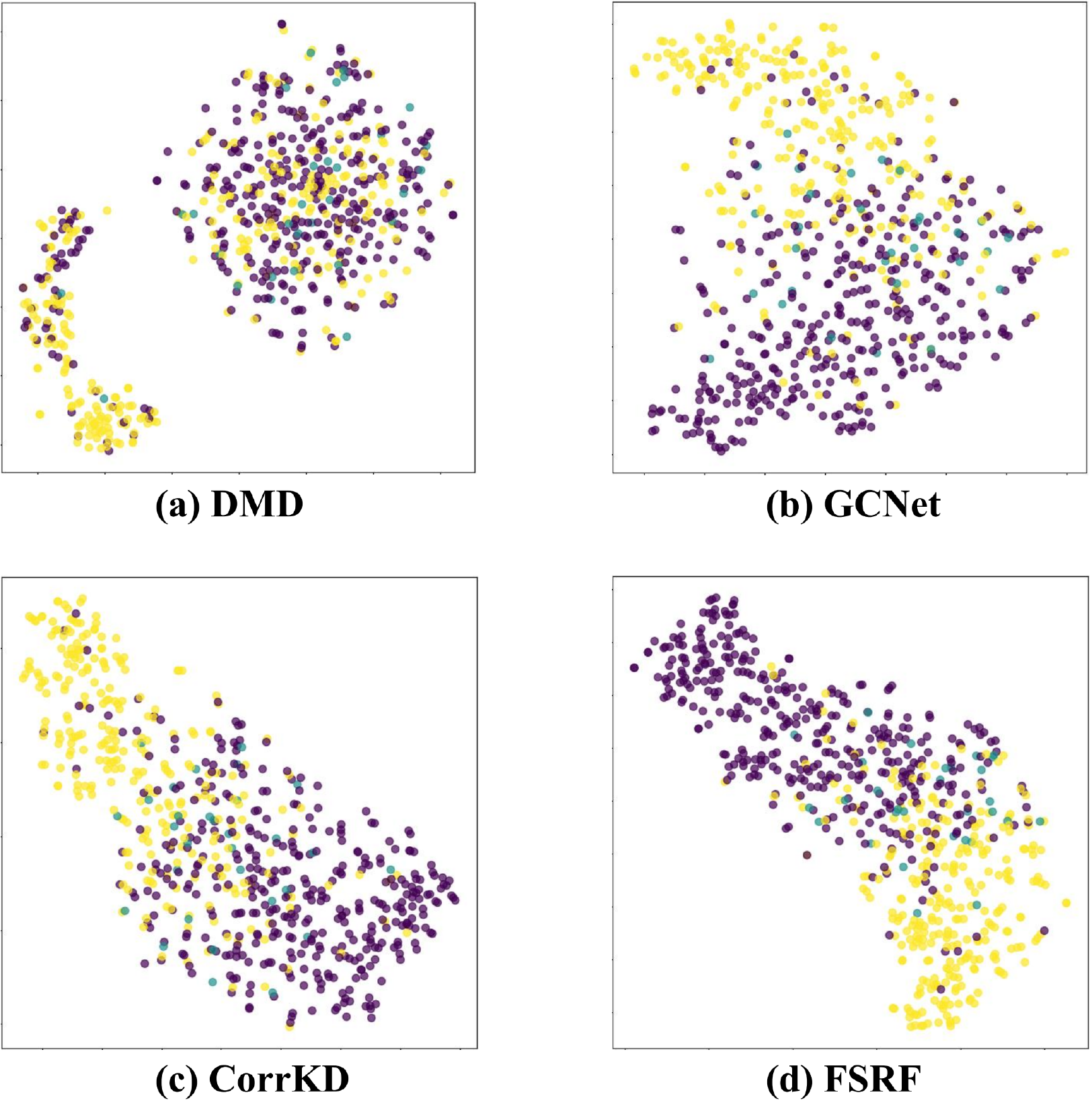}
  \caption{
 Representation visualization  on the 
MOSI dataset.
 }
  \label{vis1}
\end{figure}


We validate the effects of proposed components in FSRF, including DHF and DAS.
As illustrated in \Cref{table_ablation_FSRF}, we conducted comprehensive ablation experiments of the two missing-modality cases in the MOSI and MOSEI datasets.
The following observation proves the necessity of components:
\textbf{(\romannumeral1)} 
First, we remove the DHF and concatenate the three original modalities in the DHF. The consistent performance degradation phenomenon in both missing modality cases shows that the proposed DHF can adequately capture the critical information and sentiment semantics in modalities.
\textbf{(\romannumeral2)} Then, when our DAS is eliminated, the worse performance demonstrates that performing bidirectional distributional alignment using the self-distillation paradigm can generate favorable joint representations.

\subsection{Qualitative Analysis}
Figures \ref{vis1}(a)\,-\ref{vis1}(c) show the visualization results of baseline models and \Cref{vis1}(d) presents the visualization results of our FSRF. 
The complete modality-based baseline model (\emph{i.e.}, DMD \cite{li2023decoupled}) cannot handle the problem of modality missingness, leading to extremely confusing latent distributions.
The missing modality-based baseline models (\emph{i.e.}, GCNet \cite{lian2023gcnet}, and CorrKD \cite{li2024correlation}) mitigate the negative effects of modality missing to some extent and slightly distinguish the different categories.
In contrast, the proposed method clearly separates the potential representations of samples with different sentiments and thus has the strongest performance and robustness.

\section{Conclusion}
In this paper, we propose a Factorization-guided Semantic Recovery Framework (FSRF) to tackle various challenges posed by missing modalities in the MSA task.
Specifically, we propose a de-redundant homo-heterogeneous factorization module that factorizes modality into modality-homogeneous, modality-heterogeneous, and noisy representations and design elaborate constraint paradigms for representation learning.
Furthermore, we introduce a distribution-aligned self-distillation module that leverages bidirectional knowledge transfer under consistency constraint paradigms to effectively recover missing semantics.
Comprehensive experiments validate the superiority of FSRF.

\section*{Acknowledgment}
This work was supported in part by the National Natural Science Foundation of China (62072321), the Science and Technology Program of Jiangsu Province (BZ2024062), the Natural Science Foundation of the Jiangsu Higher Education Institutions of China  (22KJA520007), Suzhou Planning Project of Science and Technology (2023ss03).



\bibliographystyle{IEEEbib}
\bibliography{icme2025references}

\vspace{12pt}
\color{red}

\end{document}